\documentclass[10pt,twocolumn,letterpaper]{article}

\usepackage[pagenumbers]{cvpr} % To force page numbers, e.g. for an arXiv version

\usepackage{graphicx}
\usepackage{float}  
\usepackage{subcaption} 
\usepackage{booktabs}
\usepackage{multirow}
\definecolor{cvprblue}{rgb}{0.21,0.49,0.74}
\usepackage[pagebackref,breaklinks,colorlinks,allcolors=cvprblue]{hyperref}

\def\paperID{2609} % *** Enter the Paper ID here
\def\confName{CVPR}
\def\confYear{2025}

\title{IP-MOT: Instance Prompt Learning for Cross-Domain Multi-Object Tracking}

\author{Run Luo\\
SIAT\\
{\tt\small }
\and
Zikai Song\\
HUST\\
{\tt\small }
\and
Longze Chen\\
SIAT\\
{\tt\small }
\and
Yunshui Li\\
SIAT\\
{\tt\small }
\and
Min Yang\\
SIAT\\
{\tt\small }
\and
Wei Yang\\
HUST\\
{\tt\small }
}

\begin{document}
\maketitle
\begin{abstract}
Multi-Object Tracking (MOT) aims to associate multiple objects across video frames and is a challenging vision task due to inherent complexities in the tracking environment. Most existing approaches train and track within a single domain, resulting in a lack of cross-domain generalizability to data from other domains.
While several works have introduced natural language representation to bridge the domain gap in visual tracking, these textual descriptions often provide too high-level a view and fail to distinguish various instances within the same class.
In this paper, we address this limitation by developing IP-MOT, an end-to-end transformer model for MOT that operates without concrete textual descriptions. Our approach is underpinned by two key innovations: Firstly, leveraging a pre-trained vision-language model, we obtain instance-level pseudo textual descriptions via prompt-tuning, which are invariant across different tracking scenes; Secondly, we introduce a query-balanced strategy, augmented by knowledge distillation, to further boost the generalization capabilities of our model.
Extensive experiments conducted on three widely used MOT benchmarks, including MOT17, MOT20, and DanceTrack, demonstrate that our approach not only achieves competitive performance on same-domain data compared to state-of-the-art models but also significantly improves the performance of query-based trackers by large margins for cross-domain inputs.
\end{abstract}    
\section{Introduction}
\label{sec:intro}

Multi-object Tracking is one of the fundamental visual tracking tasks~\cite{CTTrack,CSWinTT,ByteTrack}, with applications ranging from human-computer interaction, surveillance, autonomous driving, etc. It aims at detecting the bounding box of the target and associating the same target across consecutive frames in a video sequence. 
Recent MOT approaches can be categorized into tracking-by-detection methods and joint detection and tracking-by-query methods. Tracking-by-detection methods have emerged as the dominant tracking paradigm for several years, powered by advances in deep learning and real-time object detectors ~\cite{YOLOX,DETR}. In this paradigm, a detector first identifies objects' bounding boxes within each single frame, followed by an association model that generates trajectories by linking these identified objects across subsequent frames. This process employs techniques such as motion-based tracking using the Kalman filter~\cite{KalmanFilter} and Re-identification (Re-ID)~\cite{chen2018real,bergmann2019tracking} methods to ensure accurate object matching~\cite{SORT,OCSORT,ByteTrack,BoTSORT, p3aformer,DeepSORT,FairMOT}. 
On the other hand, tracking-by-query methods have recently gained traction, offering a more holistic, end-to-end MOT approach. These query-based methods ~\cite{MOTR,MeMOT,TrackFormer, TransTrack} perform detection and tracking concurrently by leveraging the interplay and progressive decoding of detect and track queries within a Transformer framework.

\begin{figure}[t]
  \centering \includegraphics[width=0.98\columnwidth,keepaspectratio]{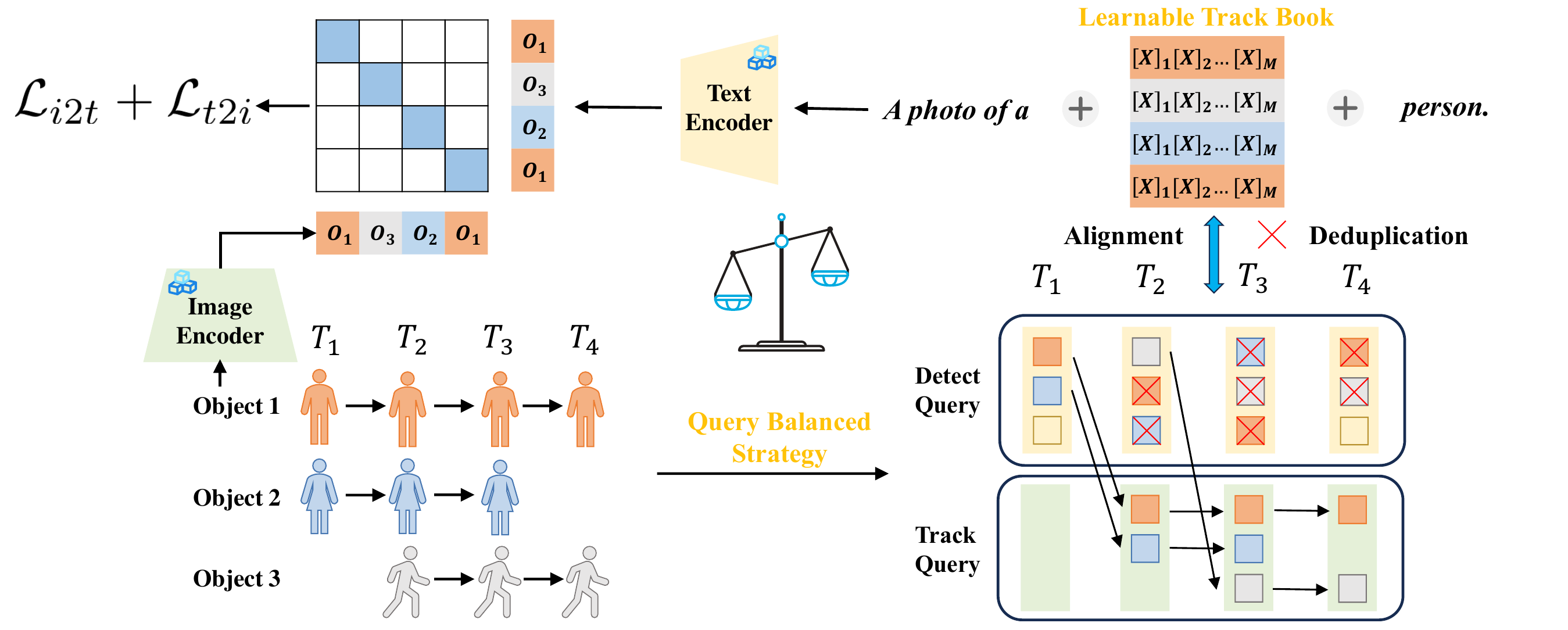}
  \caption{\textbf{IP-MOT}. We propose IP-MOT, which further improves the generalization ability of the model by using a online learnable TrackBook instead of a manually designed TrackBook to obtain a more fine-grained instance-level textual description. Meanwhile, a query balanced strategy (QBS) is also proposed to further improve the tracking and detection accuracy of IP-MOT for cross-domain and some-domain inputs.}
  \label{fig:overview}
\end{figure}
While previous methods have shown significant performance in certain contexts, their predominant focus on homogenous domains constrains their versatility. This specialization results in limited applicability across diverse scenarios, creating a development bottleneck in the MOT field. Some works ~\cite{Ltrack,OvTrack} try to integrate natural language representations to enhance domain adaptability. However, the limited availability of detailed textual descriptions has restricted the improvement of these models' generalization abilities.

Based on the analysis above, in this paper, we focus on leveraging natural language presentation by proposing a \textbf{I}nstance-level \textbf{P}rompt-learning \textbf{M}ulti-\textbf{O}bject \textbf{T}racking method with Transformer, coined as \textbf{IP-MOT}. We use pre-trained vision-language models like CLIP ~\cite{CLIP} to introduce the natural language representation into MOT models, as illustrated in Figure \ref{fig:overview}. Unlike using a hand-crafted textual descriptions, our model maintains a trainable TrackBook to generate instance-level textual descriptions for each tracked object through prompt-tuning, which contain invariant information of tracked targets across different tracking scenes. Afterward, we align output embedding with its corresponding stable textual representation through contrastive learning to improve the generalization ability. Besides, we apply triplet loss to produce a more distinguishable representation.

Although the MOTR~\cite{MOTR} architecture is elegant, it suffers from the optimization conflict between detection and association critically, which finally results in poor detection precision. Therefore, to overcome the unfair label assignment problem between detect queries and track queries, we propose a query balanced strategy where the detect queries are responsible for detecting all appeared targets and extra deduplication module is used to filter out the same target from detection results. This strategy not only refines the tracking accuracy of IP-MOT but also enriches the training dataset for textual description alignment, thereby boosting cross-domain generalization.

To evaluate the generalization performance of our model, we train our models on MOT17 and validate it on MOT20 dataset. We also evaluate our model on MOT17 and challenging DanceTrack dataset to show the performance for same-domian inputs. The experimental results reveal that our approach not only achieves competitive performance on same-domain data compared to state-of-the-art models but also significantly improves the performance of query-based trackers by large margins for cross-domain inputs. In addition, we perform extensive ablation studies to further demonstrate the effectiveness of our designs.
\section{Related Work}
\label{sec:related}

Existing MOT algorithms can be divided into two mainstream approaches according to the paradigm of handling the detection and association, \ie, the tracking-by-detection and tracking-by-query methods. 

\textbf{Tracking-by-Detection} is a common practice in the MOT field, where object detection and data association are treated as separate modules. The methods ~\cite{SORT, DeepSORT, ByteTrack, BoTSORT, OCSORT,compact,autogenic} use an existing detector~\cite{FasterRCNN, CenterNet, YOLOX} and then
% and the data association module can be further divided into motion-based methods\cite{SORT, DeepSORT, ByteTrack, BoTSORT, OCSORT} and graph-based \cite{zhang2008global,jiang2019graph,braso2020learning,li2020graph, GMTracker} methods.
%
integrate detections through a distinct motion tracker across consecutive frames, employing various techniques.
SORT ~\cite{SORT} initiated the use of the Kalman filter ~\cite{KalmanFilter} for object tracking, associating each bounding box with the highest overlap through the Hungarian algorithm ~\cite{hungarian}. DeepSORT ~\cite{DeepSORT} enhanced this by incorporating both motion and deep appearance features, while StrongSORT~\cite{StrongSORT} further integrated lightweight, appearance-free algorithms for detection and association. ByteTrack~\cite{ByteTrack} addressed fragmented trajectories and missing detections by utilizing low-confidence detection similarities.
P3AFormer~\cite{p3aformer} combined pixel-wise distribution architecture with Kalman filter to refine object association, and OC-SORT~\cite{OCSORT} amended the linear motion assumption within the Kalman Filter for superior adaptability to occlusion and non-linear motion. 

\textbf{Tracking-by-Query methods.}
In recent years, there have been several explorations into the one-stage paradigm, which combines object detection and data association into a single pipeline. Unlike the tracking-by-detection paradigm mentioned above, tracking-by-query methods apply the track query to decode the location of tracked objects progressively.
%
% \textit{Tracking-by-Query methods}, a burgeoning trend, utilize DETR \cite{DETR, DeformableDETR} extensions for MOT by representing each object as a query regressed across various frames.
Inspired by DETR-family~\cite{DETR,DeformableDETR}, most of these methods~\cite{MOTR,TrackFormer,trackssm} leverage the learnable object query to perform newborn object detection, while the track query localizes the position of tracked object. Techniques such as TrackFormer ~\cite{TrackFormer} and MOTR ~\cite{MOTR} perform simultaneous object detection and association using concatenated object and track queries. TransTrack ~\cite{TransTrack} employs cyclical feature passing to aggregate embeddings, while MeMOT ~\cite{MeMOT} encodes historical observations to preserve extensive spatio-temporal memory.

\textbf{Pre-trained Vision-Language Models}. Recently, the Pre-trained vision-language CLIP model~\cite{CLIP} measures the similarity between images and text, mapping images and their corresponding textual descriptions into a shared embedding space that allows the model to perform various tasks, such as image segmentation~\cite{denseclip,deem}, few-shot learning~\cite{few-shot} and image caption~\cite{clipcap}. In addition to the above applications, the pre-trained CLIP encoders are also applied to MOT, e.g., open-vocabulary tracking ~\cite{OvTrack}, language-guided tracking ~\cite{Ltrack} and so on. However, to the best of our knowledge, we are the first to use prompt tuning to distill the knowledge contained in the CLIP and obtain instance-level pseudo textual descriptions that can be used to boost the generalization performance of visual MOT models.

\begin{figure*}[t]
  \centering 
  \includegraphics[width=0.98\linewidth,keepaspectratio]{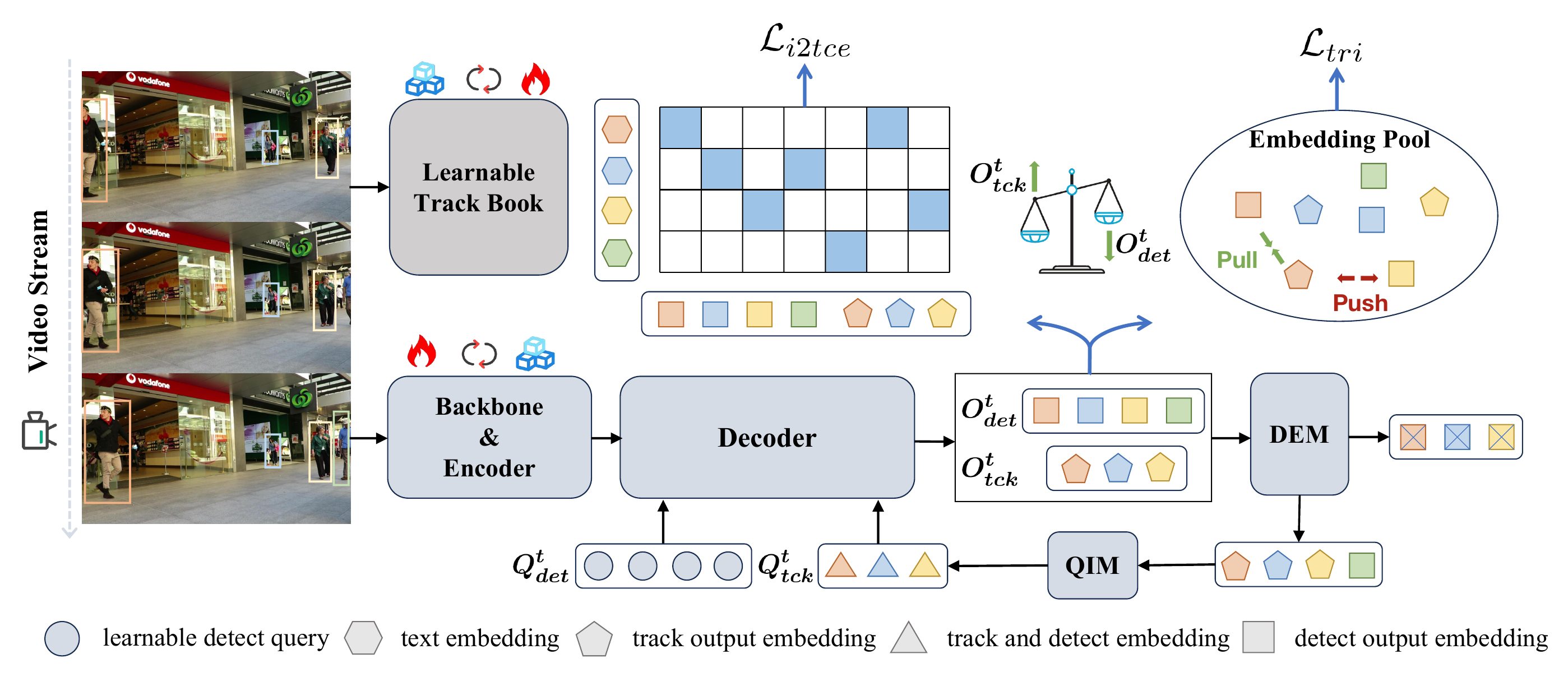}
  \caption{\textbf{The overall architecture of IP-MOT}. We use different colors to indicate different tracked targets, and the same color represents the same target. In each iteration, we first optimize our trainable TrackBook to obtain a instance-level textual description based on the target in a clip of video stream. Then, adopt a ResNet-50 ~\cite{ResNet} backbone and a Transformer ~\cite{Transformer}
  Encoder to learn a 2D representation of an input image. Afterward, the Decoder processes the detect query $Q_{det}$ and track $Q_{tck}$ , and generates the detect output embedding $O_{det}$ and track output embedding $O_{tck}$, respectively. Finally, we add output embedding into the clip-level embedding pool and align them with the corresponding frozen textual description presentation. Since the query balanced strategy (QBS) is used to alleviate unfair label assignment conflict, we designe a simple and elegant deduplication module (DEM) to duplicate detection results.}
  \label{fig:architecture}
\end{figure*}

\textbf{Domain Generalization for MOT.} 
Although the performance of the aforementioned methods
is competitive, they do not perform consistently with that of the training domain. Some works try to bridge this gap
by introducing natural language representation, such as LTrack ~\cite{Ltrack} employing hand-crafted TrackBook to inject language information into MOT, OVTrack simply adopting a constant textual description. While previous methods alleviate the generalization problem in MOT, there is still room for improvement when it comes to textual description generation strategy. Inspired by the recent advances in natural language process (NLP), we use CLIP to automatically generate distinguishable instance-level textual description via prompt tuning. In order to make better use of invariant information, we propose query balanced strategy to enhance the tracked object feature alignment over time for a more distinguishable and stable representation. Our comprehensive ablation studies further validate the effectiveness of our approach.

\section{Methodology}

\subsection{Method Overview}
We propose the \textbf{IP-MOT}, an instance-level prompt-learning Transformer for cross-domain multi-object tracking. Different from most existing methods ~\cite{Ltrack,OvTrack}, which only explicitly utilize hand-crafted textual description,
our core contribution involves constructing a TrackBook (in
Section \ref{sec:trainable}) that maintains learnable instance-level textual description for each tracked target. Additionally, we introduce a deduplication module (DEM) that effectively boosts the performance of IP-MOT via filtering out the redundant target from detection results.

As shown in Figure \ref{fig:architecture}, we use a ResNet50 ~\cite{ResNet} backbone and a Transformer Encoder to produce the image feature of an input frame $I_{t}$. Afterward by querying the encoded image feature with $[Q^{t}_{det},Q^{t}_{tck}]$, the Transformer Decoder produces the corresponding output $[O^{t}_{det},O^{t}_{tck}]$. It is worth noting that in our paper, $Q^{t}_{det}$ is responsible for detecting all targets, while $Q^{t}_{tck}$ is responsible for detecting each tracked target. Then, we predict the classification confidence $c^{t}_{i}$, bounding box $b^{t}_{i}$, and instance embedding $e^{t}_{i}$ corresponding to the $i^{th}$ target from the output emdeddings. For simplicity, we skip the training of the trainable TrackBook and directly obtain the corresponding textual representation $p^{t}_{i}$. Finally, after aligning the text representation $p^{t}_{i}$ and the instance embedding $e^{t}_{i}$, we filter out the redundant target detected by $Q^{t}_{det}$ based on the deduplication confidence $d^{t}_{i}$, output of the DEM, to retain the newborn target. The details of our components will be elaborated in the following sections.

\subsection{Trainable TrackBook}
\label{sec:trainable}
We first briefly review CLIP. It consists of two encoders, an image encoder $\mathcal{I}(\cdot)$ and a text encoder $\mathcal{T}(\cdot)$. The text encoder $\mathcal{T}(\cdot)$ and image encoder $\mathcal{I}(\cdot)$ are implemented as two transformers, which are used to generate a representation from a textual description and image respectively. Specifically, $i\in \{1...B\}$ denotes the index of the image-text pair within a batch. Let $img_i$ be the embedding of image feature, while $text_i$ is the corresponding embedding of text feature, then compute the similarity between $img_i$ and $text_i$:

\begin{equation}\label{eq:svt}
s(V_i,T_i) = V_i \cdot T_i = g_V(img_i)\cdot g_T(text_i)
\end{equation}
where $g_V(\cdot)$ and  $g_T(\cdot)$ are linear layers projecting embedding into a shared embedding space. The image-to-text contrastive loss $\mathcal{L}_{i2t}$ is calculated as:
\begin{equation}\label{eq:li2t}
\mathcal{L}_{i2t}(i) = - \log\frac{\exp(s(V_i,T_i))}{\sum_{a=1}^B \exp(s(V_i,T_a))}
\end{equation}
and the text-to-image contrastive loss $\mathcal{L}_{t2i}$:
\begin{equation}\label{eq:lt2i}
\mathcal{L}_{t2i}(i) = - \log\frac{\exp(s(V_i,T_i))}{\sum_{a=1}^B \exp(s(V_a,T_i))}
\end{equation}
% \begin{equation}\label{eq:lt2i}
% \mathcal{L}_{t2i}(i) = - \log\frac{\exp(s(V_i,T_i))}{\sum_{a=1}^B \exp(s(V_a,T_i))}
% \end{equation}
where numerators in \cref{eq:li2t} and \cref{eq:lt2i} are the similarities of two embeddings from matched pair, and the denominators are all similarities with respect to anchor $V_i$ or $T_i$.

We build trainable TrackBook by introducing ID-specific learnable tokens to learn ambiguous textual descriptions, which are independent for each object ID. Specifically, the text descriptions fed into $\mathcal{T}(\cdot)$ are designed as ``A photo of a $\rm[X]_1[X]_2[X]_3...[X]_M$ person", where each $\rm[X]_m$($\rm m\in{1,...M}$) is a learnable text token with the same dimension as word embedding. $\rm M$ indicates the number of learnable text tokens. During training phrase, we fix the parameters of $\mathcal{I}(\cdot)$ and $\mathcal{T}(\cdot)$, and only tokens $\rm[X]_m$ are optimized. 

Similar to CLIP, we use $\mathcal{L}_{i2t}$ and $\mathcal{L}_{t2i}$, but replace $text_i$ with $text_{o_{i}}$ in \cref{eq:svt}, since each object ID shares the same text description. Moreover, for $\mathcal{L}_ {t2i}$, different images in a batch probably belong to the same person, so $T_{o_{i}}$ may have more than one positive, we change it to: 

\begin{equation}\label{eq:lt2i_P}
\mathcal{L}_{t2i}(o_{i}) = -\frac{1}{|P(o_{i})|} \log\frac{\sum_{p\in P(o_{i})}\exp(s(V_p,T_{o_{i}}))}{\sum_{a=1}^B \exp(s(V_a,T_{o_{i}}))}
\end{equation}
\noindent where $P(o_{i})=\{p\in {1...B}:o_{p} = o_{i}\}$ is the set of indices of all positives for $T_{o_{i}}$ in the batch, and $|\cdot| $ is its cardinality. By minimizing the loss of $\mathcal{L}_{i2t}$ and $\mathcal{L}_{t2i}$, we can obtain textual description for each tracked object.

\subsection{Query Balanced Strategy}
Although the MOTR~\cite{MOTR} architecture is elegant, it suffers from the optimization conflict between detection and association critically. During training, the number of label assignments of the track queries is several times that of the detect queries label assignment, and the insufficient training of the detect query eventually leads to poor detection accuracy. Therefore, to overcome the unfair label assignment problem between detect queries and track queries, we propose a query balanced strategy in which detect queries are responsible for detecting all appeared targets, so that the number of both queries is balanced during supervised training. Moreover, query balanced strategy can provide more training samples for invariant textual description alignment to further boost cross-domain generalization capability. This simple but effective strategy can alleviate aforementioned problem and improve the performance of the model for same-domain and cross-domain inputs.

\begin{figure}[t]
  \centering \includegraphics[width=0.98\columnwidth,keepaspectratio]{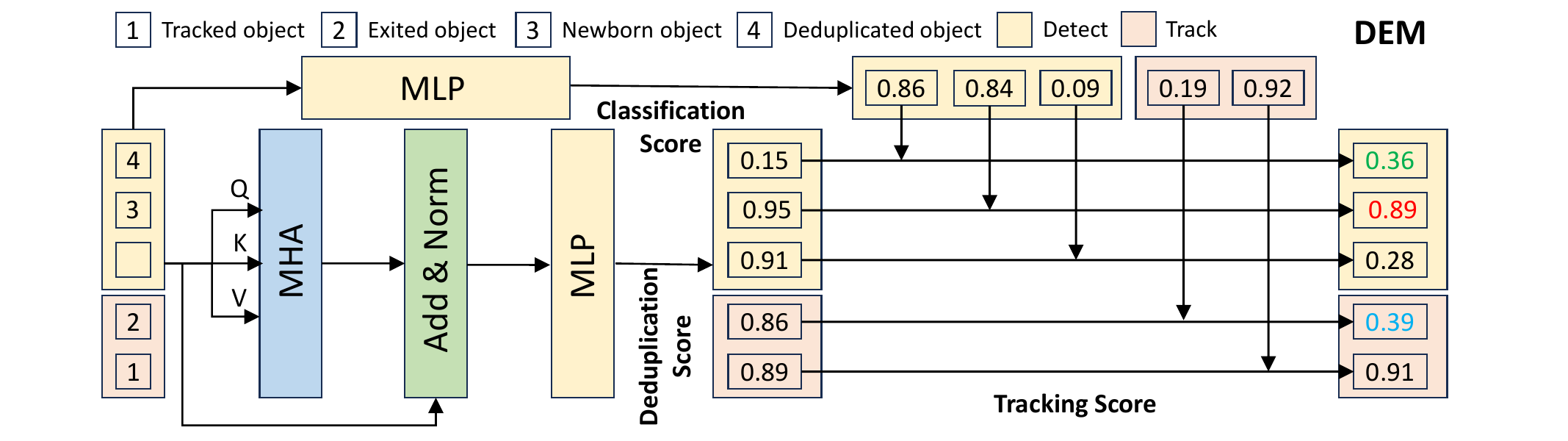}
  \caption{\textbf{The structure of deduplication module}. In the inference stage, we only restrain deduplicated objects by calculating the geometric mean of the classification score and the deduplication score to obtain the tracking score. Then in the subsequent QIM module, we keep newborn objects and drop exited objects based on the tracking score.}
  \label{fig:dem}
\end{figure}

\subsection{Deduplication Module}
Since the query balanced strategy is adopted, we design a deduplication module (DEM) which consists of a one-layer multi-head self-attention and a two-layer MLP to filter out the redundant targets in the detect query and leave only the newborn targets. As shown in Figure \ref{fig:dem}, the inputs of DEM are the hidden state $[O^{t}_{det},O^{t}_{tck}]$ produced by Transformer decoder and the corresponding
classification confidence $c^{t}=[c^{t}_{det},c^{t}_{tck}] = MLP([O^{t}_{det},O^{t}_{tck}])$. 
Afterward, we obtain deduplication confidence as $d^{t} =[d^{t}_{det},d^{t}_{tck}]=MLP(MHA([O^{t}_{det},O^{t}_{tck}]))$ and then restrain deduplicated objects by calculating tracking score as $s^{t} = \sqrt{Sigmoid(c^{t})\cdot Sigmoid(d^{t})}$. Finally in the subsequent QIM module, we keep newborn objects and drop exited objects based on the tracking score to ensure the end-to-end elegance of the approach. 

\subsection{Model Training}

For each iteration in one epoch, we first optimize our trainable TrackBook by minimizing $\mathcal{L}_{t2i}$ and $\mathcal{L}_{i2t}$ loss. In order to better train the online TrackBook, we use distributed training operation to gather all training samples located on different nodes. Afterward, we freeze the TrackBook and optimize IP-MOT with extension of the collective average loss. Given a clip $V_{\xi}$ of $N$ frames as input, the results predicted by the model are denoted as $\widehat{P}=\{\hat{p}_{i}\}_{i=1}^{N}$, and the corresponding ground-truths are $P=\{p_{i}\}_{i=1}^{N}$. The collective average loss $\mathcal{L}_{clip}$ is computed based on $\widehat{P}$ and $P$. It consists of two parts, the tracking loss and detection loss. These two losses exactly share the same form. The difference is that the tracking loss is for localizing the targets that have been recognized in previous frames, and the detection loss is to tackle the newborn targets. Mathematically, The original collective average loss $\mathcal{L}_{clip}$ can be formulated as follows:
\begin{equation}
  \begin{split}
 \hspace{-3mm}\mathcal{L}_{clip}\hspace{-0.5mm}=\hspace{-0.5mm} \frac{1}{T}\sum\limits_{n=1}^{N}\hspace{-0.5mm}(\mathcal{L}(\widehat{P}^{i}_{tck}|_{q_{t}},P^{i}_{tck})\hspace{-0.8mm}+\hspace{-0.8mm}\mathcal{L}(\widehat{P}^{i}_{det}|_{q_{d}},P^{i}_{det})\hspace{-0.5mm})
  \end{split}
  \label{Eq5}
\end{equation}
where $\widehat{P}^{i}_{tck}|_{q_{t}}$, ${P}^{i}_{tck}$, $\widehat{P}^{i}_{det}|_{q_{d}}$, and ${P}^{i}_{det}$ are the association predictions, association labels, detection predictions, and detection labels, respectively. $T$ denotes the total number of the targets in the clip $V_{\xi}$ of $N$ frames. $\mathcal{L}(\cdot)$ is implemented similarly to the one in DETR, which is formulated as:
\begin{equation}
  \begin{split}
    \mathcal{L}(\widehat{P}_{i}|_{q_{i}},P_{i}) = \lambda_{cls}\mathcal{L}_{cls} + \lambda_{l_{1}}\mathcal{L}_{l_{1}} + \lambda_{giou}\mathcal{L}_{giou}
  \end{split}
  \label{Eq6}
\end{equation}
 $\mathcal{L}_{cls}$,  $\mathcal{L}_{l_{1}}$, and $\mathcal{L}_{giou}$ are the focal loss for classification, $L_{1}$ loss for regressing width and height, and the common generalized IoU loss. $\lambda_{cls}$, $\lambda_{l_1}$, $\lambda_{giou}$ are three hyper-parameters.

We employ the triplet loss $\mathcal{L}_{tri}$ and image to text cross-entropy loss $\mathcal{L}_{i2tce}$ with label smoothing to extend collective average loss $\mathcal{L}_{clip}$ , they are calculated as:
\begin{equation}\label{eq:ltri}
\mathcal{L}_{tri}(i) = \max(\Vert e_{i}-P(e_{i})\Vert_{2})-\Vert e_{i}-N(e_{i})\Vert_{2}+\alpha,0)
\end{equation}
\begin{equation}\label{eq:li2tce}
\mathcal{L}_{i2tce}(i) = \sum_{k=1}^T -q_k\log\frac{\exp(s(V_i,T_{o_k}))}{\sum_{{o_a}=1}^T \exp(s(V_i,T_{o_a}))}
\end{equation}
\noindent here $q_k = (1-\epsilon)\delta_{k,y}+\epsilon/T$ denotes value in the target distribution, $\Vert e_{i}-P(e_{i})\Vert_{2}$ and $\Vert e_{i}-N(e_{i})\Vert_{2}$ are $l_2$ euclidean distance of positive pair and negative pair, while $\alpha$ is the margin of $\mathcal{L}_{tri}$. Eventually, The extended collective average loss $\mathcal{L}_{clip}^{*}$ can be formulated as follows: 
\begin{equation}
  \begin{split}
\hspace{-3mm}\mathcal{L}_{clip}^{*}\hspace{-0.5mm}=\hspace{-0.5mm} \mathcal{L}_{clip}+ \frac{1}{T}\sum\limits_{n=1}^{T}\hspace{-0.5mm}(\lambda_{tri} \mathcal{L}_{tri}\hspace{-0.8mm}+\hspace{-0.8mm}\lambda_{i2tce}\mathcal{L}_{i2tce}\hspace{-0.5mm})
  \end{split}
  \label{Eq5:new}
\end{equation}

\section{Experiments}
\subsection{Datasets and Metrics}
 \noindent \textbf{Datasets}. We evaluate our method on multiple multi-object tracking datasets, including MOT17 ~\cite{MOT16}, MOT20 ~\cite{MOT20} and DanceTrack ~\cite{DanceTrack}. MOT17 and MOT20 are used for pedestrian tracking, where targets mostly move linearly, while scenes in MOT20 are more crowded. To verify the domain generalization ability of models to unseen domains, we train our model on MOT17 and validate them on MOT20. Additionally, we assess the same-domain performance of IP-MOT on DanceTrack, a challenging dataset have a similar appearance, severe occlusion, and frequent crossovers with highly non-linear motion. 

\noindent \textbf{Metric}. We mainly use the higher order (HOTA ~\cite{hota}) metrics and CLEAR-MOT metrics to evaluate our method. Specifically, HOTA metrics consist of higher order tracking accuracy (HOTA), association accuracy score (AssA) and detection accuracy score (DetA). CLEAR-MOT Metrics include IDF1 score (IDF1) ~\cite{idf1}, multiple object tracking accuracy (MOTA) ~\cite{mota} and identity switches (IDS). Among them, HOTA, AssA, and IDF1 are crucial metrics for comparing tracking performance, while MOTA, DetA are the pivotal metrics for comparing detection performance.
 
\subsection{Implementation Details}
\quad We adopt the visual encoder $\mathcal{I}(\cdot)$ and the text encoder $\mathcal{T}(\cdot)$ from CLIP as the backbone to optimize trainable TrackBook. We choose the ViT-B/16, which contains 12 transformer layers with the hidden size of 768 dimensions to extract image feature. To match the output of the $\mathcal{T}(\cdot)$, the dimension of the image feature vector is reduced from 768 to 512 by a linear layer. we use the AdamW optimizer with a learning rate initialized at 3.5e-4 and decayed by a cosine schedule to optimize the learnable text tokens $\rm[X]_m$($\rm m\in{1,...M}$) in TrackBook.  For each iteration, we first resize the cropped target to 128x256 and train our learnable TrackBook. And then, we align IP-MOT with frozen instance-level textual description.

Following MeMOTR~\cite{MeMOTR}, we build IP-MOT based on DAB-Deformable-DETR ~\cite{DAB-DERT}, which is pre-trained on COCO and employs ResNet50 as backbone. To make fair comparison, we also provide the results of our model based on original Deformable-DETR ~\cite{DeformableDETR} in Table \ref{tab:dancetrack}. During the training process, the batch size is 1 and each batch contains a multi-frame video clip. The frames in each clip are selected from training videos with a random interval between 1 to 10. We use the AdamW ~\cite{adamw} optimizer with the initial learning rate of 2.0e-4. Our models are conducted on PyTorch with 8 NVIDIA GeForce RTX 3090 with the some data augmentation strategy of MeMOTR, which includes random random flip and random crop. During the training process, $\lambda_{cls}$, $\lambda_{l_{1}}$, and $\lambda{giou}$ are set as 2, 5, and 2, while $\lambda_{tri}$, $\lambda_{i2tce}$, and the margin of $\alpha$ are set as 2, 4, and 0.3. IP-MOT is trained for totally 20 epochs and the learning rate decays by 10 at the $10^{th}$ epoch on DanceTrack dataset. For MOT17, we train our model on  a joint train set with additional CrowdHuman val set ~\cite{shao2018crowdhuman} for totally 120 epochs and the learning rate decays by 10 at the $60^{th}$ epoch. For simplicity, we set score threshold $\tau = 0.5$ in our experiments.

\subsection{Same-domain State-of-the-art Comparison}

\noindent \textbf{Comparison on the DanceTrack Dataset}.
To evaluate IP-MOT under same-domain challenging
non-linear object motion, we compare IP-MOT with the state-of-the-art methods on the DanceTrack test set. As shown in Table \ref{tab:dancetrack}, IP-MOT* surpasses MOTR ~\cite{MOTR} by 7.7 (61.9 vs. 54.2) on HOTA, 6.5 (46.7 vs. 40.2) on AssA and 8.9 (60.4 vs. 51.5) on IDF1. Our method alleviates the unfair label assignment conflicts between detect and track queries by query balanced strategy. Therefore, IP-MOT* also has a significant improvement in detection accuracy, such as 4.8 (84.5 vs. 79.7) on MOTA and 2.5 (76.0 vs. 73.5) on DetA in addition to the improvement in tracking accuracy. Eventually, our method achieves 61.9 HOTA, 48.7 AssA, and 62.0 IDF1 without extra data or memory mechanism, showing promising potential by competitive performance compared with the state-of-the-art methods. 

\begin{table}[t]
  \centering
  \begin{minipage}[c]{0.48\textwidth}
    \centering
    \captionof{table}{Same-domain performance comparison to SOTA approaches on the Dancetrack test set. The best results are shown in \textbf{bold}. * means the result based on standard
Deformable-DETR ~\cite{DeformableDETR}.}
    \resizebox{1.0\linewidth}{!}{
      \begin{tabular}{l|ccccc}
    \toprule
    \multirow{2}{*}{\textbf{Methods}} & \multicolumn{5}{c}{DanceTrack (Same-domain)} \\
    % \midrule[0.5pt]
    \cmidrule(lr){2-6}
     & HOTA & MOTA & DetA & AssA & IDF1 \\
    \midrule[0.5pt]
    % \cmidrule(lr){1-6}
    \textcolor{gray}{\textit{with extra data/memory:}}\\
    % MT-IoT & 66.7 & 94.0 & 84.1 & 53.0 & 70.6 \\
    \quad\quad \textcolor{gray}{MeMOTR} ~\cite{MeMOTR} & \textcolor{gray}{68.5} & \textcolor{gray}{89.9} & \textcolor{gray}{80.5} & \textcolor{gray}{58.4}& \textcolor{gray}{71.2} \\
    \quad\quad \textcolor{gray}{MOTRv2} ~\cite{MOTRv2} & \textcolor{gray}{69.9} & \textcolor{gray}{91.9} & \textcolor{gray}{83.0} & \textcolor{gray}{59.0} & \textcolor{gray}{71.7} \\
    \midrule[0.5pt]
    % \cmidrule(lr){1-6}
    \textit{w/o extra data/memory:}\\
    \quad \quad QDTrack ~\cite{QDTrack} & 45.7 & 83.0 & 72.1 & 29.2 & 44.8 \\
    \quad \quad FairMOT ~\cite{FairMOT} & 39.7 & 82.2 & 66.7 & 23.8 & 40.8 \\
    \quad \quad TraDes ~\cite{TraDes} & 43.3 & 86.2 & 74.5 & 25.4 & 41.2 \\
    \quad \quad SORT ~\cite{SORT} & 47.9 & \textbf{91.8} & 72.0 & 31.2 & 50.8 \\
    \quad \quad ByteTrack ~\cite{ByteTrack} & 47.3 & 89.5 & 71.6 & 31.4 & 52.5 \\
    \quad \quad OC-SORT ~\cite{OCSORT} & 54.6 & 89.6 & 80.4 & 40.2 & 54.6 \\
    \quad \quad TransTrack ~\cite{TransTrack} & 45.5 & 88.4 & 75.9 & 27.5 & 45.2 \\
    \quad \quad MOTR ~\cite{MOTR} & 54.2 & 79.7 & 73.5 & 40.2 & 51.5 \\
    \quad \quad CenterTrack ~\cite{CenterTrack} & 41.8 & 86.8 & 78.1 & 22.6 & 35.7 \\
    \quad \quad GTR ~\cite{zhou2022global} & 48.0 & 84.7 & 72.5 & 31.9 & 50.3 \\
    \quad \quad DiffusionTrack ~\cite{DiffusionTrack}& 52.4 & 89.3 & \textbf{82.2} & 33.5 & 47.5 \\
    \quad \quad C-BIoU ~\cite{C-BIoU} & 60.6 & 91.6 & 81.3 & 45.4 & 61.6 \\
    \quad \quad \textbf{IP-MOT*} (ours) & 59.5 & 84.5 & 76.0 & 46.7 & 60.4 \\
    \quad \quad \textbf{IP-MOT} (ours) & \textbf{61.9} & 88.2 & 79.0 & \textbf{48.7} & \textbf{62.0} \\
    \bottomrule
      \end{tabular}
    }
    \label{tab:dancetrack}
  \end{minipage}
  % \hspace{0.02\textwidth}
  \begin{minipage}[c]{0.48\textwidth}
    \centering
    \captionof{table}{Same-domain performance comparison to state-of-the-art approaches on the MOT17 test set with the private detections. The best results are shown in \textbf{bold}.}
    \resizebox{1.0\linewidth}{!}{
      \begin{tabular}{l|cccccc}
    \toprule
    \multirow{2}{*}{\textbf{Methods}} & \multicolumn{6}{c}{MOT17 (Same-domain)} \\
    % \midrule[0.5pt]
    \cmidrule(lr){2-7}
     & MOTA & IDF1 & HOTA & AssA & DetA & IDS\\
    \midrule[0.5pt]
    \textit{CNN based:}\\
    \quad GTR ~\cite{zhou2022global} & 75.3 & 71.5 & 59.1 & 57.0 & 61.6 & /  \\
    \quad TubeTK ~\cite{pang2020tubetk} & 63.0 & 58.6 & / & / & / & 4137 \\
    \quad CenterTrack ~\cite{CenterTrack} & 67.8 & 64.7 & 52.2 & 51.0 & 53.8 & 3039 \\
    \quad ByteTrack ~\cite{ByteTrack} & 80.3 & 77.3 & 63.1 & 62.0 & 64.5 & 2196 \\
    \quad FairMOT ~\cite{FairMOT} & 73.7 & 72.3 & 59.3 & 58.0 & 60.9 & 3303 \\
    \quad StrongSORT ~\cite{StrongSORT} & 79.6 & 79.5 & 64.4 & 64.4 & 64.6 & 1194  \\
    \quad OC-SORT ~\cite{OCSORT} & 78.0 & 77.5 & 63.2 & 63.4 & 63.2 & 1950 \\
    \quad BoT-SORT ~\cite{BoTSORT} & 80.5 & 80.2 & 65.0 & 65.5 & 64.9 & 1212  \\
    \midrule[0.5pt]
    \textit{Transformer based:}\\
    \quad TrackFormer ~\cite{TrackFormer} & 74.1 & 68.0 & 57.3 & 54.1 & 60.9 & 2829 \\
    \quad TransTrack ~\cite{TransTrack} & \textbf{74.5} & 63.9 & 54.1 & 47.9 & \textbf{61.6} & 3663 \\
    \quad TransCenter ~\cite{TransCenter} & 73.2 & 62.2& 54.5 & 49.7 & 60.1 & 4614  \\
    \quad MeMOT ~\cite{MeMOT} & 72.5 & 69.0 & 56.9 & 55.2 & / & 2724 \\
    \quad MOTR ~\cite{MOTR} & 71.9 & 68.4 & 57.2 & 55.8 & 58.9 & 2115 \\
    \quad LTrack ~\cite{Ltrack} & 72.1 & 69.1 & 57.5 & 56.1 & 59.4 & 2100 \\
    % MeMOTR& 72.8 & 71.5 & 58.8 & 58.4 & 59.6 & / \\
    \quad \textbf{IP-MOT} (ours) & 73.2 & \textbf{69.6} & \textbf{58.2} & \textbf{56.4} & 60.4 & \textbf{1896} \\
    \bottomrule
    \end{tabular}
    }
    \label{tab:mot17}
  \end{minipage}
  \vspace{-12pt}
\end{table}

\noindent \textbf{Comparison on the MOT17 Dataset}. Query-based trackers suffer from serious overfitting problems in MOT17 since the number of training set in MOT17 is insufficient to train an end-to-end tracker. Therefore, our method only slightly improves the performance compared to original MOTR ~\cite{MOTR}. As illustrated in Table \ref{tab:mot17}, IP-MOT obtains the metrics 58.2 HOTA, 56.4 AssA, and 69.6 IDF1 on MOT17. The competitive results indicate that IP-MOT can tackle same-domain tracking scenes well. Notably, IP-MOT also obtains competitive performance on the detection related metrics (60.4 DetA and 73.2 MOTA). Meanwhile, it only produces 1896 IDS, which is the lowest among all compared methods. The more continuous tracklets generated by IP-MOT demonstrate that our proposed learnable TrackBook can produce more distinguishable textual representations. Compared to the MOTR ~\cite{MOTR}, the introduction of learnable text token and query balanced strategy consistently improves the detection (DetA) and association (AssA) accuracy by 1.4\% and 2.6\% correspondingly. These experimental results further validate the effectiveness of our designs.

\begin{table}[tp]\small
  \centering
    \caption{Cross-domain performance comparison to state-of-the-art approaches on the MOT20 train set with the private detections. The best results are shown in \textbf{bold}. The used datasets of all methods are marked out in the column “Data” of this table (CH and 17 refer to CrowdHuman and MOT17, respectively). Notably, all methods are trained and validated in the same setting.}
  \setlength{\tabcolsep}{4.5mm}{
    \begin{tabular}{@{ }l|c@{ }c@{ }c@{ }c@{ }c@{ }c}
    \toprule
    \multirow{2}{*}{\textbf{Methods}} & \multicolumn{6}{c}{MOT20 (Cross-domain)} \\
    \cmidrule(lr){2-7}
      & MOTA & IDF1 & HOTA & AssA & DetA & Data\\
    \midrule[0.5pt]
    CenterTrack ~\cite{CenterTrack}  & 42.9 & 39.0 & 29.7 & 25.6 & 35.0 & CH+17\\
    FairMOT ~\cite{FairMOT}  & 57.6 & 53.8 & 41.9 & 35.9 & 49.7 & CH+17\\
    TraDeS ~\cite{TraDes}  & 44.9 & 39.3 & 28.0 & 25.5 & 32.7 & CH+17\\
    CSTrack ~\cite{CSTrack} & 49.6 & 44.9 & 33.9 & 29.8 & 38.8 & CH+17\\
    OMC ~\cite{OMC} & 55.9 & 49.4 & 38.8 & 32.2 & 46.9 & CH+17\\
    MTrack ~\cite{MTrack} & 54.8 & 52.9 & 40.6 & 37.0 & 44.9 & CH+17\\
    % TrackFormer ~\cite{TrackFormer} & CH+17 & - & - & - & - & - \\
    TransTrack ~\cite{TransTrack} & 58.1 & 44.8 & 35.8 & 27.3 & 47.3 & CH+17\\
    % TransCenter ~\cite{TransCenter} & CH+17 & - & - & - & - & -  \\
    MOTR ~\cite{MOTR} & 54.2 & 56.0 & 43.1 & 42.3 & 43.9 & CH+17\\
    MeMOTR ~\cite{MeMOTR} & 55.7 & 56.4 & 43.1 & 41.8 & 44.7 & CH+17\\
    LTrack ~\cite{Ltrack} & 57.4 & 60.4 & 46.2 & 43.8 & 48.2 & CH+17\\

    \textbf{IP-MOT} (ours) & \textbf{68.3} & \textbf{62.5} & \textbf{49.2} & \textbf{44.6} & \textbf{55.3}& CH+17 \\
    \bottomrule
    \end{tabular}
  }
  \label{tab:mot20}
  \vspace{-3pt}
\end{table}

\subsection{Cross-domain State-of-the-art Comparison}

\noindent \textbf{Comparison on the MOT20 Dataset}.  
We test our approach on the cross-domain evaluation benchmark proposed by LTrack ~\cite{Ltrack}.
As presented in Tab \ref{tab:mot20}, the performance drop of IP-MOT is relatively small, while the performance of compared end-to-end methods drops significantly. Specifically, IP-MOT achieves 49.2 HOTA, 44.6 AssA and 62.5 IDF1, which significantly outperforms all compared methods by large margins. The results indicate that the generalization ability of IP-MOT to unseen domains is promising. In addition, IP-MOT surpasses previous SOTA method, LTrack, by 3.0 (49.2 vs. 46.2) on HOTA, 0.8 (44.6 vs. 43.8) on AssA and 2.1 (62.5 vs. 60.4) on IDF1, which further confirms the benefit of aliging with fine-grained instance-level textual description. We further prove our components’ effectiveness in Section \ref{sec:ablation}

\subsection{Ablation Study}
\label{sec:ablation}
In this section, we study several components of our model through ablation studies. All the
experiments are conducted on the cross-domain evaluation benchmark. To accelerate the ablation study process. We train our model on the MOT17 train-half set for totally 10 epochs and evaluate it on the MOT20 val-half set by TrackEval. 

\begin{table}[tp]
  \centering
  \begin{minipage}[c]{0.48\textwidth}
    \centering
    \captionof{table}{Ablation study of the length of learnable text token in TrackBook prompt, which is denoted as $\rm M$.}
    \resizebox{1.0\linewidth}{!}{
      \begin{tabular}{c|cccccc}
    \toprule
    $\rm M$ & HOTA $\uparrow$ & MOTA $\uparrow$ & IDF1$\uparrow$ & DetA $\uparrow$ & AssA $\uparrow$\\
    \midrule[0.5pt]
    2 & 31.4 & 38.5 & 42.7 & 29.8 & 33.8 \\
    4 & \textbf{33.1} & \textbf{40.9} & \textbf{45.7} & \textbf{31.0} & 36.0 \\
    6 & 32.0 & 38.0 & 44.6 & 28.8 & \textbf{36.7} \\
    8 & 31.0 & 37.9 & 41.7 & 29.7 & 32.8 \\
    \bottomrule
      \end{tabular}
    }
    \label{tab:prompt-length}
  \end{minipage}
  \hspace{0.02\textwidth}
  \begin{minipage}[c]{0.48\textwidth}
    \centering
    \captionof{table}{Ablation experiments on the layers of the usage of new  collective average loss in multi-layer auxiliary loss $\mathcal{L}_{clip*}$.}
    \resizebox{1.0\linewidth}{!}{
      \begin{tabular}{cc|cccc}
    \toprule
    $L_{\mathcal{L}_{clip}^{*}}$ & HOTA $\uparrow$ & MOTA $\uparrow$ & IDF1$\uparrow$ & DetA $\uparrow$ & AssA $\uparrow$\\
    \midrule[0.5pt]
    1  & 28.5 & 31.8 & 38.7 & 23.9 & 32.6 \\
    3  & 31.0 & 40.2 & 43.2 & 30.2 & 34.6 \\
    5 & \textbf{33.1} & \textbf{40.9} & \textbf{45.7} & \textbf{31.0} & \textbf{36.0} \\
    \bottomrule
      \end{tabular}
    }
    \label{tab:layers-new-loss}
  \end{minipage}
  \\
    \begin{minipage}[c]{0.48\textwidth}
    \centering
    \captionof{table}{Ablation study on the triplet loss $\mathcal{L}_{tri}$ and image to text cross entropy loss $\mathcal{L}_{i2tce}$.}
    \resizebox{1.0\linewidth}{!}{
      \begin{tabular}{cc|ccccc}
    \toprule
    $\mathcal{L}_{tri}$ & $\mathcal{L}_{i2tce}$ & HOTA $\uparrow$ & MOTA $\uparrow$ & IDF1 $\uparrow$ & AssA $\uparrow$ \\
    \midrule[0.5pt]
    - & - & 29.8 & 36.8 & 41.2 & 32.6 \\
    \checkmark & - & 32.2 & 39.8 & 44.6 & 35.0 \\
    - & \checkmark & 31.2 & 37.4 & 43.2 & 35.6 \\
    \checkmark & \checkmark & \textbf{33.1} & \textbf{40.9} & \textbf{45.7} & \textbf{36.0} \\
    \bottomrule
      \end{tabular}
    }
    \label{tab:tri-i2tce-loss}
  \end{minipage}
    \hspace{0.02\textwidth}
      \begin{minipage}[c]{0.48\textwidth}
    \centering
    \captionof{table}{Ablation study on instance-level textual representation alignment (Align) and query balanced strategy (QBS). \textit{na\"ive} meas the result based on standard Deformable-DETR ~\cite{DeformableDETR}.}
    \resizebox{1.0\linewidth}{!}{
      \begin{tabular}{cc|cccc}
    \toprule
    $Align$ & $QBS$ & HOTA $\uparrow$ & MOTA $\uparrow$ & IDF1 & AssA $\uparrow$\\
    \midrule[0.5pt]
    \multicolumn{2}{c}{\textit{na\"ive}} & 24.6 & 24.9 & 30.3 & 28.7 \\
    \midrule[0.5pt]
    - & - & 29.8 & 36.8 & 41.2 & 32.6 \\
    \checkmark & - & 33.1 & 40.9 & 45.7 & 36.0 \\
    - & \checkmark & 29.6 & 36.8 & 39.8 & 31.7 \\
    \checkmark & \checkmark & \textbf{36.9} & \textbf{48.8} & \textbf{50.2} & \textbf{36.7} \\
    \bottomrule
    \end{tabular}
    }
    \label{tab:align-qbs}
  \end{minipage}
  \vspace{-12pt}
\end{table}

\noindent \textbf{Trainable TrackBook}. The prompt in TrackBook are designed as ``A photo of a $\rm[X]_1[X]_2[X]_3...[X]_M$ person", where each $\rm[X]_m$($\rm m\in{1,...M}$) is a learnable text token with the same dimension as word embedding and $\rm M$ indicates the length of learnable text tokens. The length of learnable text tokens $\rm M$ determines the semantic richness of the textual description, which in turn affects the performance of the IP-MOT. We experimentally search for a suitable length $\rm M$, as show in Table \ref{tab:prompt-length}. Increasing $\rm M$
from 2 to 4 dramatically improves HOTA and AssA metrics by 5.4\% and 6.5\%, respectively. However, continuing to increase the length of learnable text tokens will cause semantic sparsity problems, thus slightly weakening the overall performance.

\begin{figure}[t]
    \centering
    \begin{subfigure}{0.10\textwidth}
        \centering
        \includegraphics[width=\textwidth]{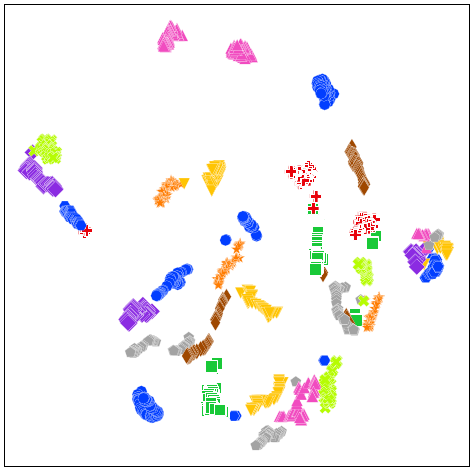}
        \caption{MOTR}
        \label{fig:vis-motr}
    \end{subfigure}%
      \hspace{0.01\textwidth}
        \begin{subfigure}{0.18\textwidth}
        \centering
        \includegraphics[width=\textwidth]{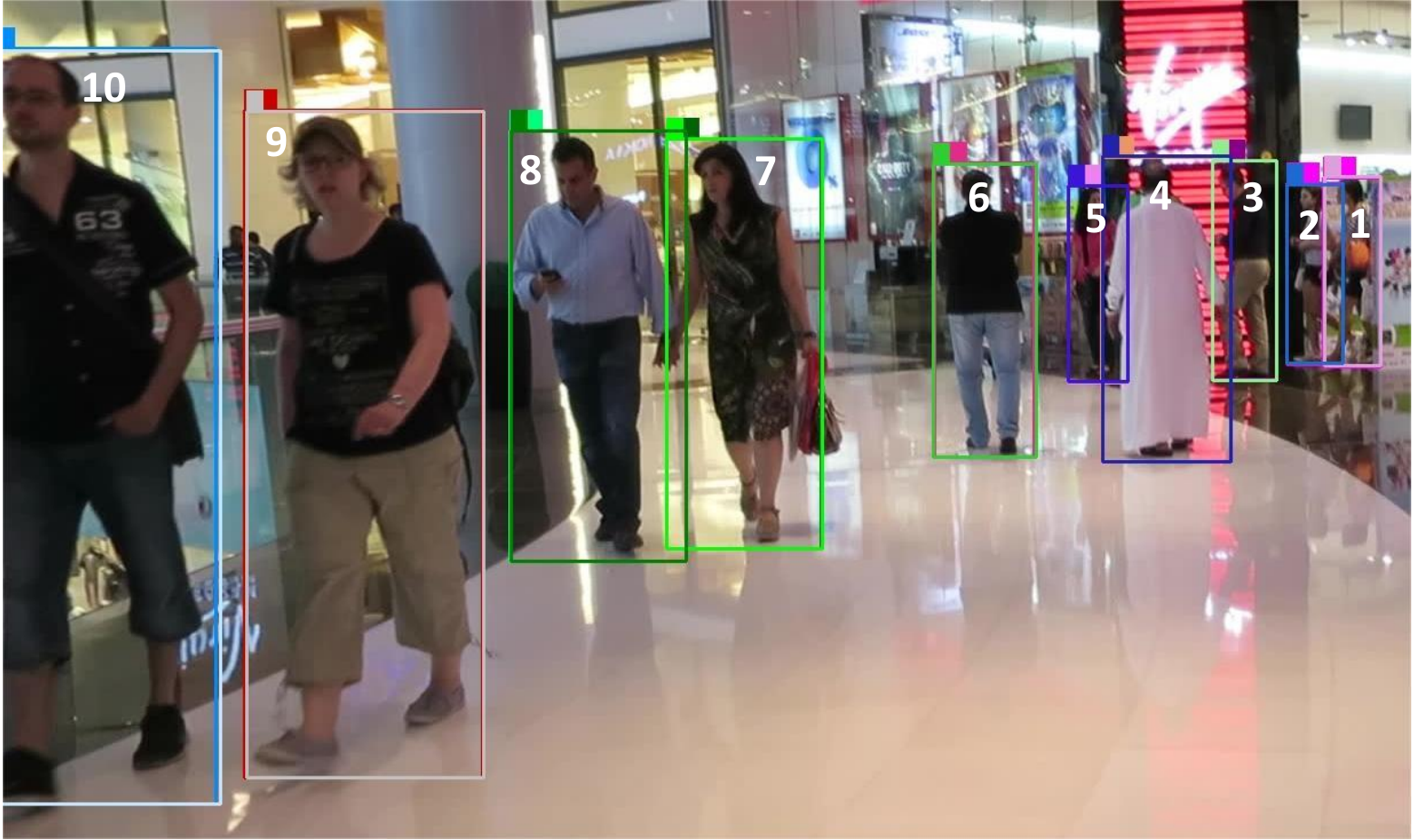}
    \caption{Query prediction}
        \label{fig:vis-dem-mot17}
    \end{subfigure}%
    \begin{subfigure}{0.13\textwidth}
        \centering
        \includegraphics[width=\textwidth]{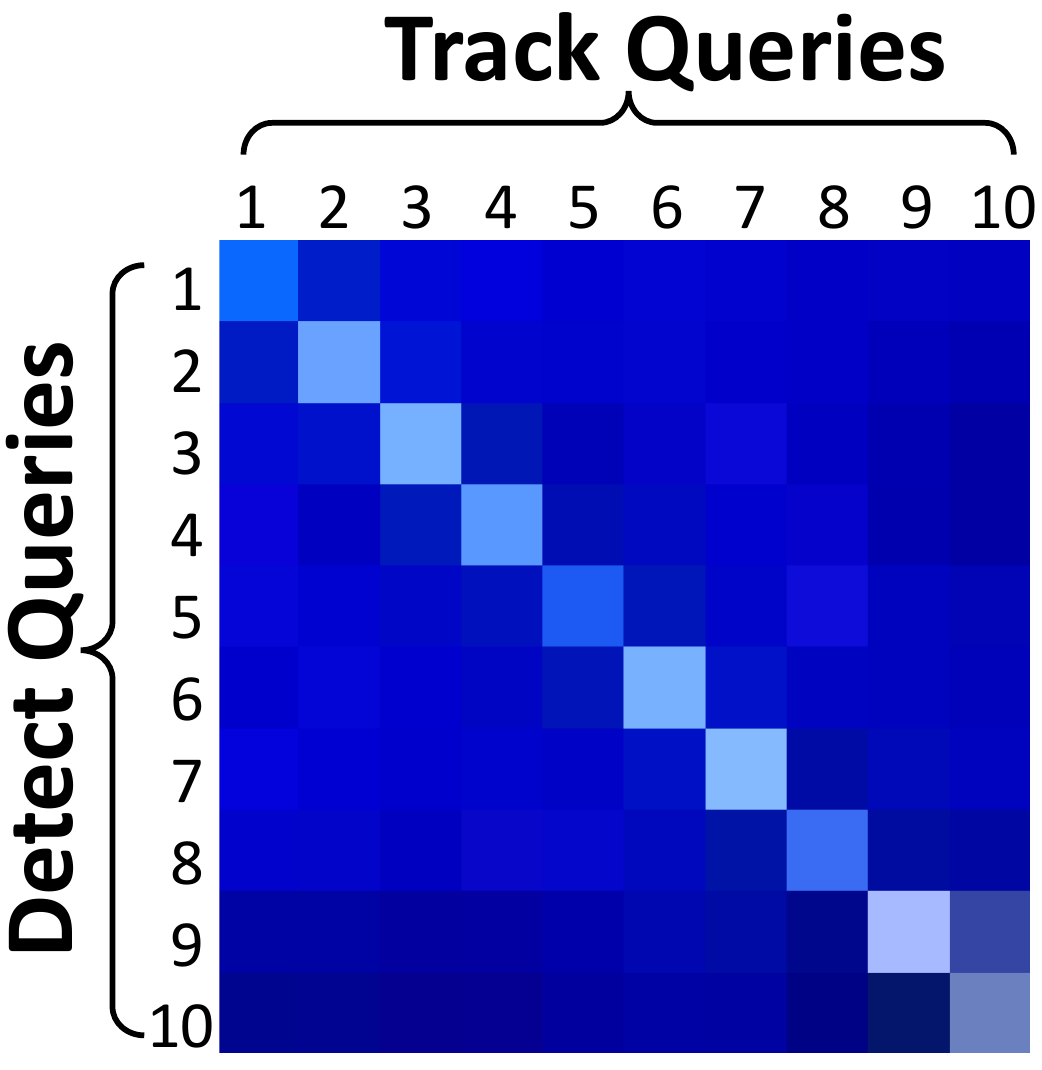}
        \caption*{Query attention}
        \label{fig:vis-dem-atten-mot17}
    \end{subfigure}
    \\
    \begin{subfigure}{0.10\textwidth}
        \centering
        \includegraphics[width=\textwidth]{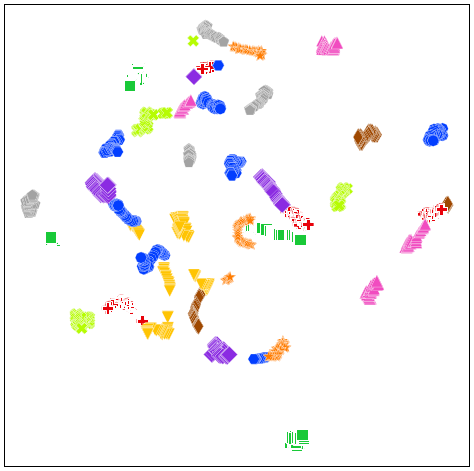}
        \caption{IP-MOT}
        \label{fig:vis-IP-MOT}
    \end{subfigure}
         \hspace{0.01\textwidth}
        \begin{subfigure}{0.18\textwidth}
        \centering
        \includegraphics[width=\textwidth]{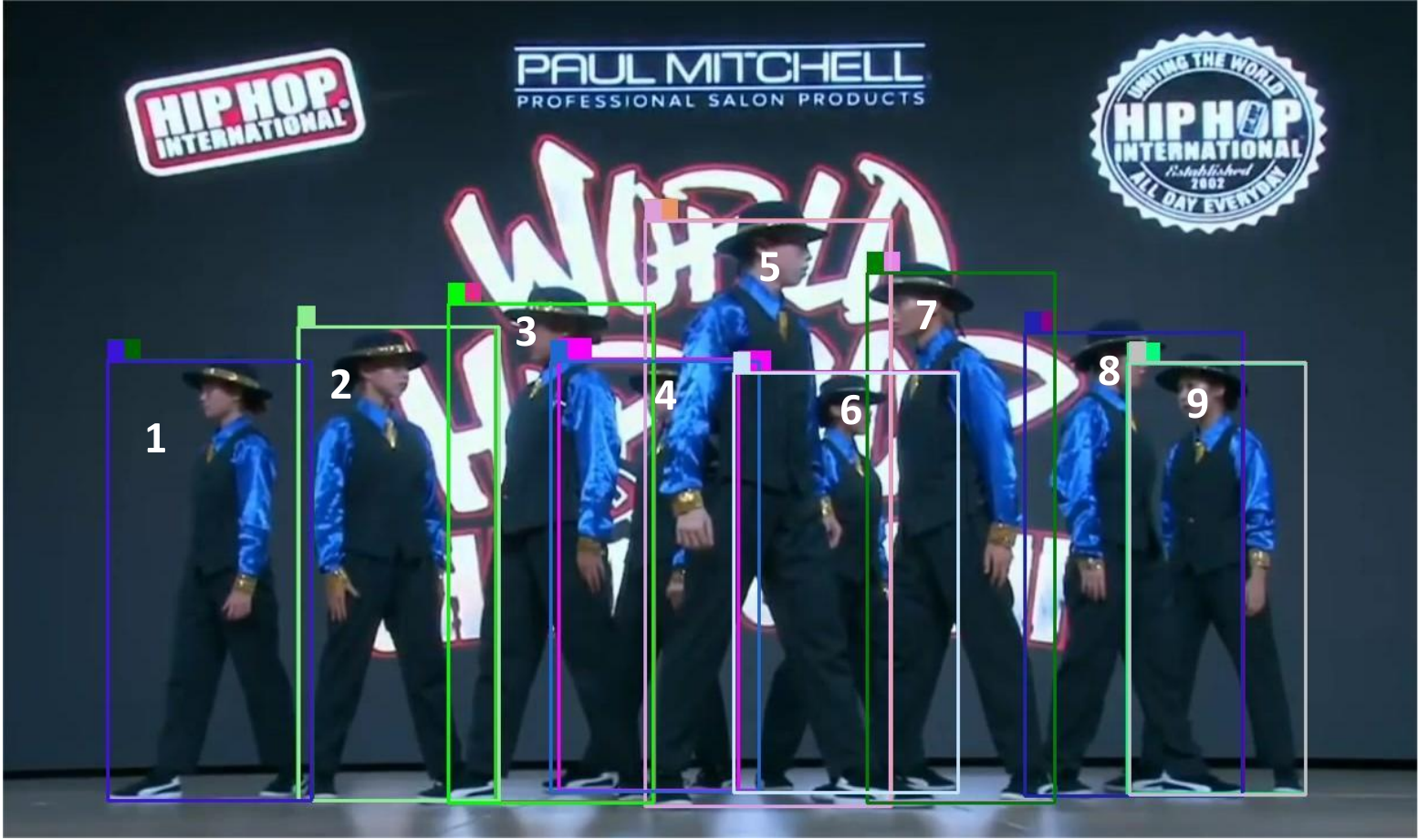}
        \caption{Query prediction}
        \label{fig:vis-dem-dance}
    \end{subfigure}%
    \begin{subfigure}{0.13\textwidth}
        \centering
        \includegraphics[width=\textwidth]{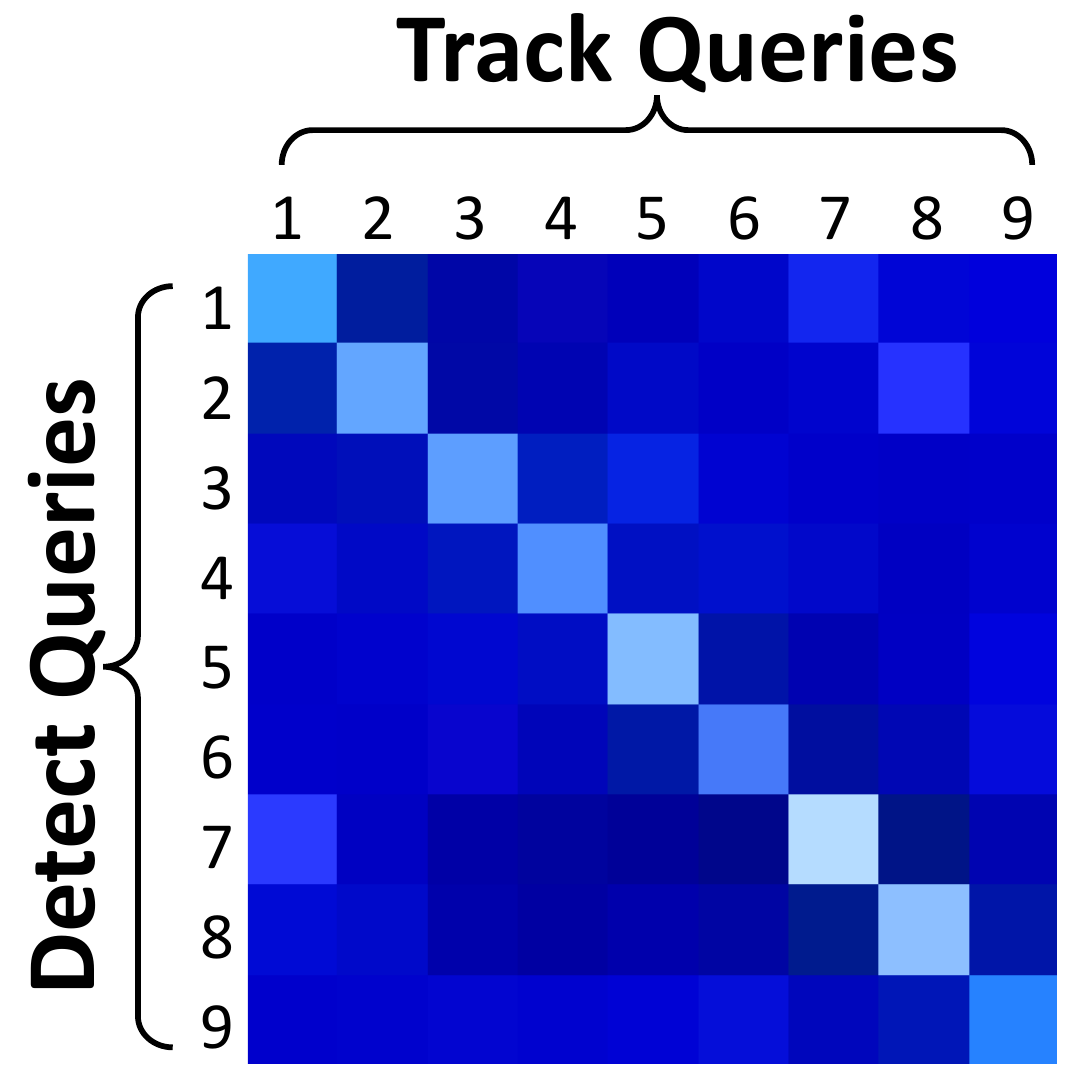}
        \caption*{Query attention}
        \label{fig:vis-dem-atten-dance}
    \end{subfigure}

    \caption{Visualization of track Output Embedding $O_{tck}$ (the first 50 frames in sequence MOT20-02 on cross-domain benchmark) by using t-Distributed Stochastic Neighbor Embedding (t-SNE). Embeddings for different targets are marked in different colors and shapes. Our method (\ref{fig:vis-IP-MOT}) helps the model learn a more stable and distinguishable representation than MOTR (\ref{fig:vis-motr}) for the track output embedding. Corresponding tracking performance is shown in Table \ref{tab:align-qbs}. Visualization of \ref{fig:vis-dem-mot17}, \ref{fig:vis-dem-dance} shows IP-MOT track query box prediction highly overlaps the detect query box prediction on the same-domain MOT17 and DanceTrack test set respectively, and the corresponding  query self-attention map shows a clear exchange of information between the dedupliacted detect query and the track query of the same instance, demonstrating the effectiveness of our DEM.}
    \label{fig:vis-contrast}
\end{figure}

\noindent \textbf{New Collective Average Loss}. We extend original collective average loss $\mathcal{L}_{clip}$ by adding triplet loss $\mathcal{L}_{tri}$ and image to text cross entropy loss $\mathcal{L}_{i2tce}$ to get new collective average loss $\mathcal{L}_{clip*}$. We explore the effect of the layers of replacing the $\mathcal{L}_{clip}$ with $\mathcal{L}_{clip*}$ in multi-layer auxiliary loss on the performance of the model. As shown in Table \ref{tab:layers-new-loss}, increasing the layers of $\mathcal{L}_{clip*}$ in multi-layer auxiliary loss can steadily improve the model's generalization performance.
$\mathcal{L}_{i2tce}$ is responsible for aligning the target embedding with the corresponding text description representation to make the target association more stable. $\mathcal{L}_{tri}$ try to make target embedding more distinguishable by shortening the distance between same targets and alienating the distance between different targets. In Table \ref{tab:tri-i2tce-loss}
, our experimental results show that using either of these two losses can improve the generalization ability of the model, and the performance can be further enhanced by using them together. Therefore, the $\mathcal{L}_{clip*}$ loss helps the model learn a more stable and distinguishable representation, as visualized in Figure \ref{fig:vis-motr} and \ref{fig:vis-IP-MOT}. 

\begin{figure*}[tp]
    \centering
    \begin{subfigure}{0.11\textwidth}
        \centering
        \includegraphics[width=\textwidth,keepaspectratio,page=1]{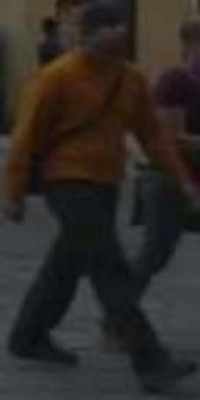}
        \caption*{target* \#1}
        \label{fig:target1}
    \end{subfigure}%
    \hfill
        \begin{subfigure}{0.11\textwidth}
        \centering
        \includegraphics[width=\textwidth,keepaspectratio,page=5]{figs/vis.pdf}
        \caption*{target \#1}
        \label{fig:s-target1}
    \end{subfigure}%
    \hfill
        \begin{subfigure}{0.11\textwidth}
        \centering
        \includegraphics[width=\textwidth,keepaspectratio,page=9]{figs/vis.pdf}
        \caption*{dog \#1}
        \label{fig:s-color1}
    \end{subfigure}%
    \hfill
        \begin{subfigure}{0.11\textwidth}
        \centering
        \includegraphics[width=\textwidth,keepaspectratio,page=13]{figs/vis.pdf}
        \caption*{cat \#1}
        \label{fig:s-texture1}
    \end{subfigure}%
    \hfill    
    \begin{subfigure}{0.11\textwidth}
        \centering
        \includegraphics[width=\textwidth,keepaspectratio,page=2]{figs/vis.pdf}
        \caption*{target* \#2}
        \label{fig:target2}
    \end{subfigure}%
    \hfill
        \begin{subfigure}{0.11\textwidth}
        \centering
        \includegraphics[width=\textwidth,keepaspectratio,page=6]{figs/vis.pdf}
        \caption*{target \#2}
        \label{fig:s-target2}
    \end{subfigure}%
    \hfill
        \begin{subfigure}{0.11\textwidth}
        \centering
        \includegraphics[width=\textwidth,keepaspectratio,page=10]{figs/vis.pdf}
        \caption*{dog \#2}
        \label{fig:s-color2}
    \end{subfigure}%
    \hfill
        \begin{subfigure}{0.11\textwidth}
        \centering
        \includegraphics[width=\textwidth,keepaspectratio,page=14]{figs/vis.pdf}
        \caption*{cat \#2}
        \label{fig:s-texture2}
    \end{subfigure}%
    \\
    \begin{subfigure}{0.11\textwidth}
        \centering
        \includegraphics[width=\textwidth,keepaspectratio,page=3]{figs/vis.pdf}
        \caption*{target* \#3}
        \label{fig:target3}
    \end{subfigure}%
    \hfill
        \begin{subfigure}{0.11\textwidth}
        \centering
        \includegraphics[width=\textwidth,keepaspectratio,page=7]{figs/vis.pdf}
        \caption*{target \#3}
        \label{fig:s-target3}
    \end{subfigure}%
    \hfill
        \begin{subfigure}{0.11\textwidth}
        \centering
        \includegraphics[width=\textwidth,keepaspectratio,page=11]{figs/vis.pdf}
        \caption*{dog \#3}
        \label{fig:s-color3}
    \end{subfigure}%
    \hfill
        \begin{subfigure}{0.11\textwidth}
        \centering
        \includegraphics[width=\textwidth,keepaspectratio,page=15]{figs/vis.pdf}
        \caption*{cat \#3}
        \label{fig:s-texture3}
    \end{subfigure}%
    \hfill
        \begin{subfigure}{0.11\textwidth}
        \centering
        \includegraphics[width=\textwidth,keepaspectratio,page=4]{figs/vis.pdf}
        \caption*{target* \#4}
        \label{fig:target4}
    \end{subfigure}%
    \hfill
        \begin{subfigure}{0.11\textwidth}
        \centering
        \includegraphics[width=\textwidth,keepaspectratio,page=8]{figs/vis.pdf}
        \caption*{target \#4}
        \label{fig:s-target4}
    \end{subfigure}%
    \hfill
        \begin{subfigure}{0.11\textwidth}
        \centering
        \includegraphics[width=\textwidth,keepaspectratio,page=12]{figs/vis.pdf}
        \caption*{dog \#4}
        \label{fig:s-color4}
    \end{subfigure}%
    \hfill
        \begin{subfigure}{0.11\textwidth}
        \centering
        \includegraphics[width=\textwidth,keepaspectratio,page=16]{figs/vis.pdf}
        \caption*{cat \#4}
        \label{fig:s-texture4}
    \end{subfigure}%

    \caption{\textbf{Visualization of instance-level textual description}. Since the Stable Diffuion ~\cite{StableDiffusion} and the CLIP ~\cite{CLIP} share a same text encoder, we can generate the corresponding image based on the instance-level textual description. Target* means the original target in MOT17 dataset, while target, color, and texture means corresponding synthetic image by replacing the last word in textual description with ``person",``dog", and ``cat", respectively.}
    
    \label{fig:vis-semantic}
    \vspace{-12pt}
\end{figure*}

\noindent \textbf{Query Balanced Strategy}. We perform the alignment by using the new collective average loss $\mathcal{L}_{clip*}$ and employ query balanced strategy (QBS) to further enhance the performance. In addition to explore the new collective average loss $\mathcal{L}_{clip*}$, we also ablate QBS in Table \ref{tab:align-qbs}. It shows that by using the new collective average loss with QBS, our IP-MOT achieves much better performance (36.9 vs. 33.1 on HOTA), especially improving MOTA by 19.3\%. However, without alignment, QBS produces worse performance (-3.4\% IDF1 and -2.8\% AssA). We explain that $\mathcal{L}_{clip*}$ and QBS can complement each other and ultimately achieve a better performance. Specifically, QBS can provide more training samples for clip-level embedding pool for alignment training, while alignment training can provide QBS with more stable and distinguishable out embedding to alleviate deduplication difficulties, as visualized in Figure \ref{fig:vis-dem-mot17} and \ref{fig:vis-dem-dance}. Therefore, using them together can obtain enhanced performance.

\subsection{Visualization}
To better demonstrate the superiority of IP-MOT, we visualize some instance-level textual description from TrackBook. Leveraging the shared text encoder between the Stable Diffusion model ~\cite{StableDiffusion} and CLIP ~\cite{CLIP}, we generated corresponding images based on these detailed textual descriptions. Figure \ref{fig:vis-semantic} displays various targets from the training set alongside their synthetic counterparts depicting person, dog, and cat. These synthesized images accurately capture the original targets' attributes, including color, texture, and high-level semantic information like clothing, hats, bags, and gender. In contrast, LTrack ~\cite{Ltrack}, the previous cross-domain SOTA method, utilized a manually designed TrackBook, limiting the embedding's interpretability and semantic richness. Our method overcomes these constraints by efficiently generating textual descriptions with robust semantics, stability, and recognizability, thus enhancing the model's generalizability.

\vspace{-12pt}
\section{Conclusion}

In this paper, we propose IP-MOT, an end-to-end instance-level prompt-learning Transformer for cross-domain multi-object tracking without concrete textual description. Our method builds a trainable TrackBook to obtain stable textual description for each tracked object and exploits this description to augment the representation of track embedding, thereby improving cross-domain association performance. Furthermore, through the use of a query balanced strategy, our model improves the detection accuracy, making various targets more distinguishable and stable. Consequently, IP-MOT not only exhibits competitive performance on same-domain MOT benchmarks, but also achieves the state-of-the-art performance on cross-domain MOT benchmarks. Comprehensive ablation experiments and visualizations substantiate the effectiveness of our components. We hope that future work will pay more attention on leveraging textual descriptions in multi-object tracking.

\noindent \textbf{Limitation}. While IP-MOT demonstrates excellent performance in cross-domain MOT benchmarks, it faces challenges in scenarios with multiple similar targets in the same frame. This difficulty stems from the limitations of distinguishing targets using only textual and semantic information. In such cases, appearance cues are unreliable, necessitating spatial priors or advanced post-processing for effective tracking.
{
    \small
    \bibliographystyle{ieeenat_fullname}
    \bibliography{main}
}


\end{document}